\documentclass[letterpaper, 10 pt, conference]{ieeeconf}  % Comment this line out if you need a4paper

\IEEEoverridecommandlockouts                              % This command is only needed if 
\title{\LARGE \bf
A Hybrid Optimization Framework for Grasp Synthesis under Partial Observations
}

\author{Wenzheng Zhang$^{1,*}$, Fahira Afzal Maken$^{2}$, Tin Lai$^{1}$, Fabio Ramos$^{1,3}$% <-this % stops a space
\thanks{$^*$ Corresponding author: wzha2981@uni.sydney.edu.au}
	\thanks{$^1$ School of Computer Science, The University of Sydney, Australia}
        \thanks{$^2$ Data61, CSIRO, Australia}
	\thanks{$^3$ NVIDIA, USA}
}

\usepackage[short]{optidef}
\usepackage{amsmath}
\usepackage{amssymb}
\let\labelindent\relax
\usepackage{enumitem}
\RequirePackage[activate={true,nocompatibility},final,tracking=true,kerning=true,spacing=true,factor=1100,stretch=10,shrink=10]{microtype}
\microtypecontext{spacing=nonfrench}\linepenalty=1000
\usepackage{mathptmx}
\usepackage{graphicx}
\usepackage{caption}
\graphicspath{{Images/}}
\usepackage[ruled, shortend, linesnumbered]{algorithm2e}

\usepackage{bm}
\usepackage{amssymb}
\usepackage{multirow}
\usepackage{siunitx}
\usepackage{booktabs}
\usepackage[font=small]{caption}
\usepackage{lmodern}
\usepackage{comment}
\usepackage[pagebackref=false,breaklinks=true,colorlinks,bookmarks=false]{hyperref}
\newlength{\textfloatsepsave} 

\usepackage[left=0.75in, right=0.75in, top=0.81in, bottom=0.75in]{geometry}
\usepackage[T1]{fontenc}
\usepackage{lmodern}

\begin{document}

\maketitle
\thispagestyle{empty}
\pagestyle{empty}
\setlength{\textfloatsep}{10pt plus 1.0pt minus 0.5pt}

%%%%%%%%%%%%%%%%%%%%%%%%%%%%%%%%%%%%%%%%%%%%%%%%%%%%%%%%%%%%%%%%%%%%%%%%%%%%%%%%
\begin{abstract}
\small
We propose a hybrid grasp synthesis framework that combines a learning-based Energy-Based Model (EBM) with an analytical Iterative Closest Point (ICP) method to generate robust grasps from partially observed point clouds. The learned energy function acts as a prior within a Stein Variational Gradient Descent (SVGD) framework, guiding iterative refinement of grasp configurations. Evaluated on 67 objects with 5,360 grasp attempts, our method achieves an average success rate of 60.9\%, outperforming AnyGrasp (31.1\%) and Grasp Pose Detection (48.4\%) and AS-ICP (56.6\%). These results highlight the strong generalization ability of our approach and demonstrate how combining data-driven learning with geometric optimization addresses the limitations of either strategy in isolation.

\end{abstract}

%%%%%%%%%%%%%%%%%%%%%%%%%%%%%%%%%%%%%%%%%%%%%%%%%%%%%%%%%%%%%%%%%%%%%%%%%%%%%%%%
\section{INTRODUCTION}

Grasp synthesis remains a challenging problem in robotic manipulation, particularly for unseen objects under partial observation. Traditional methods relying on complete object models often fail to generalize to the diversity of everyday objects~\cite{analytic_drawback}. To address this, we propose a hybrid grasp synthesis framework that integrates a data-driven Energy-Based Model (EBM)~\cite{Contrast} with Iterative Closest Point (ICP)~\cite{ICP} within the Stein Variational Gradient Descent (SVGD) framework~\cite{SVGD}.

Point cloud completion is one strategy for handling partial observations, but generating accurate shapes remains difficult~\cite{SplitPSO}. Although some methods generate grasps directly from partial point clouds~\cite{AnyGrasp}, they still require full object models to build training datasets~\cite{GPD}. Annealed Stein ICP (AS-ICP)~\cite{AS-ICP} is an optimization-based approach capable of generating grasps from partial inputs. However, the method is sensitive to gripper aperture, requires multiple preshapes to ensure coverage.

To overcome these limitations, we propose a hybrid approach that combines learned global priors with geometric optimization. First, an EBM is trained on data generated by AS-ICP, assigning low energy to successful grasps and high energy to failures. Integrating the EBM gradient into SVGD enables iterative refinement of grasp poses, blending learned grasp-quality cues with local geometric alignment to improve robustness.

We evaluated our approach on 57 Google Scanned Objects and 10 KIT Dataset objects, comprising 5360 grasp attempts. Our method achieves an average success rate of 60.9\%, outperforming baselines such as AnyGrasp (31.1\%), GPD (48.4\%), and AS-ICP (56.6\%).

\textbf{Contributions:}
\begin{itemize}
\item A hybrid framework that integrates EBMs and ICP within SVGD for robust grasp synthesis from partial point clouds.
\item A comprehensive ablation study analyzing model architecture, loss functions, and dataset design.
\item Extensive experiments demonstrating improved robustness, repeatability, and generalization compared to analytic, data-driven, and weakly integrated hybrid baselines.
\end{itemize}

\section{RELATED WORK}\label{sec:related}
Grasp synthesis approaches can be broadly categorized into optimization-based and data-driven methods~\cite{Survey}.

Optimization-based methods formulate grasp planning as an explicit optimization problem over full-object models. Examples include CPO-PPO~\cite{CPO-PPO}, ISF~\cite{ISF}, MDISF~\cite{MDISF}, and PPO-JPO~\cite{PPO-JPO}, which jointly optimize gripper pose and finger joints. AS-ICP~\cite{AS-ICP} simplifies the problem by optimizing only the grasp pose, reducing complexity while maintaining robustness.

Data-driven methods learn grasp quality directly from sensory data such as RGB-D or point clouds. Dex-Net~\cite{Dex-Net} employs multiview CNNs, but still relies on full object models for training. GraspNet~\cite{GraspNet}, AnyGrasp~\cite{AnyGrasp}, and GPD~\cite{GPD} predict grasps from partial point clouds, though datasets are typically generated using full models. Recent work~\cite{full_data_1} follows this paradigm, often refining candidates with ICP~\cite{CenterGrasp}.

Hybrid methods aim to combine analytical reasoning with learning, but most are weakly integrated. Typical strategies include the use of analytical quality metrics during training~\cite{DexNet2}, as evaluation criteria~\cite{GPD}, or as post-processing~\cite{ContactNet}.

Finally, Energy-Based Models (EBMs)~\cite{Contrast} have shown promise for grasp synthesis due to their flexibility in modeling distributions. However, challenges include intractable partition functions~\cite{partition} and highly non-convex energy landscapes~\cite{EBM_2}. Stein Variational Gradient Descent (SVGD)~\cite{SVGD} provides an efficient inference strategy and has been successfully applied in robotics, including trajectory ~\cite{SMP} and motion planning~\cite{steindif}~\cite{steinQ}, and localization~\cite{steinpf}.

We integrate EBMs with ICP, combining learned priors with geometric optimization under the SVGD framework, targeting robust grasp synthesis from partial point clouds.

\section{Preliminaries} \label{sec:Preliminaries}
\subsection{Energy Based Model}
Energy-Based Models (EBMs) offer a flexible framework to represent complex probability distributions. In an EBM, each input $x$ is assigned a scalar energy $E_\theta(x)$, where lower energy values indicate more plausible or desirable configurations. The probability density is defined as:

\begin{equation}
p_\theta(x) = \frac{\exp(-E_\theta(x))}{Z_\theta},
\end{equation}
where $E_\theta(x)$ is a non-linear energy function such as a deep neural network parameterized by $\theta$, and the partition function $Z_\theta$ is given by:

\begin{equation}
Z_\theta = \int \exp(-E_\theta(x)) \, dx.
\end{equation}
Inference with EBMs involves finding the configuration $x$ that minimizes the energy function~\cite{Contrast}:

\begin{equation}
\hat{x} = \arg\min_x E_\theta(x).
\end{equation}

Training EBMs is challenging due to the intractability of the partition function. Various techniques—such as contrastive divergence ~\cite{CD} and score matching~\cite{ScoreMatching}—have been proposed to address this difficulty. In our work, we focus on integrating the gradient of the EBM into the Stein Variational Gradient Descent (SVGD) framework to optimize grasp synthesis from partial point cloud data. We used different weights to balance the gradients from EBM and SGD-ICP.

\subsection{Stein Variational Gradient Descent}
Stein Variational Gradient Descent (SVGD)~\cite{SVGD} is a particle-based variational inference method that approximates a complex target distribution using a set of interacting particles. Unlike traditional variational inference methods that assume a fixed parametric form for the approximate posterior, SVGD represents the target distribution non-parametrically as an empirical distribution over particles.

In the SVGD framework, the \(j\)th particle, \(\theta_j\), is updated as follows~\cite{SVGD}:
\begin{equation}
\theta_j \leftarrow \theta_j + \eta\, \hat{\phi}^*(\theta_j), \label{eq:svgd-update}
\end{equation}
where \(\eta\) is the step size and the optimal update direction \(\hat{\phi}^*(\theta)\) is given by:
\begin{equation}
\hat{\phi}^*(\theta) = \frac{1}{K}\sum_{j=1}^{K}\left[\, k(\theta_j,\theta)\,\nabla_{\theta_j}\log p(\theta_j) + \nabla_{\theta_j} k(\theta_j,\theta)\right]. \label{eq:svgd}
\end{equation}
Here, \(k(\cdot,\cdot)\) is a positive definite kernel that couples the particles. The first term can be seen as an attractive force that pulls the particles toward regions of high probability density, and the second term acts as a repulsive force that prevents the particles from collapsing to a single mode, thereby encouraging diversity in the particle set.

\section{Stein Energy-Based Grasp Synthesis}\label{sec:method}

\subsection{Problem Formulation}
In this work, we address the problem of grasp synthesis for unknown objects using a hybrid optimization framework that integrates data-driven and analytical approaches. Given the gripper’s inner surface point cloud, $\mathcal{S}$, and the object’s point cloud captured from a single viewpoint, $\mathcal{R}$, our goal is to determine an optimal grasp pose, parameterized by $\theta = (t,q)$, where $t = {x, y, z}$ represents the 3D translation and $q ={q_w, q_x, q_y, q_z}$ represents the unit quaternion (with $q_w$ being the scalar and ${q_x, q_y, q_z}$ the vector part). Initial poses are uniformly sampled and refined using Stein Variational Gradient Descent (SVGD) to minimize a hybrid cost function, combining geometric alignment and learned grasp-quality assessments.

\begin{figure*}[t]
    \centering
    \small
    \includegraphics[width=14cm]{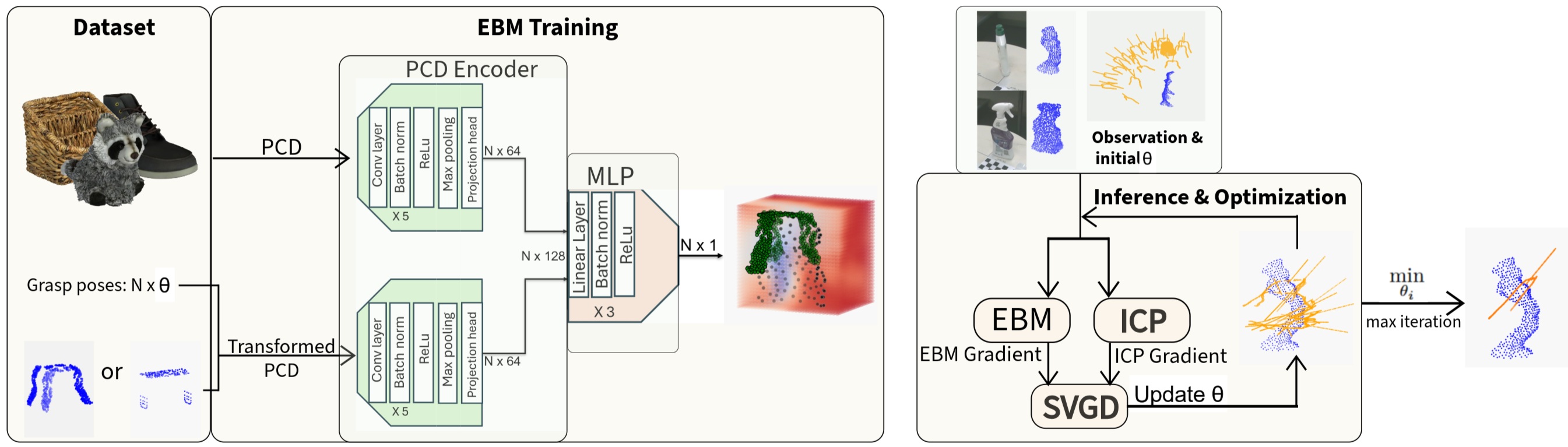}

    \caption{A brief summary of our algorithm. Both the object and gripper point clouds are fed into separate point cloud encoders to produce 64-dimensional features. These features are then concatenated and scored by the EBM to output a single energy value. The system uses SVGD to iteratively update the transformation parameters $\theta$ by leveraging gradients from the EBM and the cost function of ICP. The best pose is selected with minimum energy and matching error.}
    \label{fig:alg}
\end{figure*}

Formally, we define our grasp optimization as follows:

\begin{equation}
\begin{aligned}
&\min_{\theta} \quad \mathcal{L}(y,x) + E(y,x)\
&s.t. \quad dist(y, SDF(x)) > 0,
\label{eq:c5:icp-objective}
\end{aligned}
\end{equation}
where $x = T_{\theta}(\mathcal{S})$ denotes the transformed gripper surface point cloud under pose $\theta$, and $y = \mathcal{R}$ denotes the object's observed point cloud. Here, $\mathcal{L}(y,x)$ represents the geometric loss computed via Iterative Closest Point (ICP) matching~\cite{sgdicp}. $E(y,x)$ is the learned energy term, computed from an Energy-Based Model (EBM) trained on grasp outcomes (success/failure) evaluated in simulation, reflecting learned probabilistic knowledge of grasp quality. The constraint ensures that the resulting grasp pose is physically plausible and collision-free, enforced by evaluating the distance from the object points to the gripper's Signed Distance Field (SDF). A summary of our method is provided in Figure~\ref{fig:alg}.

\subsection{Data Generation}\label{sec:Data}
To build our dataset, we use the AS-ICP algorithm~\cite{AS-ICP} on single-view point clouds of objects. Specifically, we select 57 distinct objects from the Google Scanned Objects dataset~\cite{GoogleScan} and 10 distinct objects from the KIT dataset ~\cite{KIT}, capture eight different point clouds per object with four different orientations (0, 90, 180 and 270 degree), each with two camera elevations (0.1 and 0.7 m). For each point cloud, AS-ICP is applied to optimize collision-free grasp poses. These optimized poses are subsequently validated in the Isaac Gym simulator~\cite{Isaac}. A grasp is considered successful if the gripper maintains a firm hold on the object even after it is subjected to shaking. This process yields a labeled dataset of grasp poses. 

\subsection{Network Architecture}\label{sec:EBM}

Our EBM is designed to capture the interaction between the object and the gripper by processing their point clouds and the associated grasp pose information. The model consists of three primary modules:

\subsubsection{Point Cloud Encoder}
We adopt a PointNet-based encoder~\cite{PointNet}, as point clouds are unordered sets of 3D points without a regular grid structure. PointNet handles this by using permutation-invariant operations, making it well-suited for extracting robust geometric features for grasping. The Point Cloud Encoder module applies five one-dimensional convolutional layers (with kernel size 1), each followed by batch normalization and ReLU activations. After these convolutions, a global max pooling operation aggregates features across all points to produce a fixed-dimensional embedding. This embedding is further projected into a lower-dimensional feature space via a fully connected layer, layer normalization, and a ReLU activation.

\subsubsection{Energy Function Network}
The features are then fed into the Energy Function network, which is implemented as a multi-layer perceptron. This network comprises three fully-connected layers (interleaved with batch normalization and ReLU activations) and culminates in a single linear output that represents the energy. 

\subsection{Network Training}\label{sec:Training}
\subsubsection{Loss Function}
We employ a contrastive loss function to train our EBM for grasp synthesis. The objective is to assign low energy values to positive grasp samples and high energy values to generated negative samples. We use a smooth, differentiable variant of the hinge loss as follows~\cite{Contrast}:

\begin{equation}
L = \log \left( 1 + \exp \left( E^+ - E^- \right) \right).
\label{eq:c5:training_loss}
\end{equation}
Here, $E^+$ represents the energy of a positive sample, while $E^-$ denotes the energy of a corresponding negative sample. In the context of grasping, even a small deviation in the positive grasp pose may lead to failure, resulting in many subtle negative samples. Thus, preserving this specific pairing is crucial for effective model performance.

\subsubsection{Dataset}\label{sec:TrainingData}
The dataset generated in Section \ref{sec:Data} is highly imbalanced, containing significantly more failure cases than successful grasp poses. This imbalance can negatively impact the performance of our EBM training. In Section \ref{sec:ablation}, we discuss various data-processing strategies to mitigate such imbalances.

For training, we selected 50 objects. Given that most objects are non-symmetrical, each point cloud captured under different orientations and elevations is treated as a distinct group. From each group, we randomly select 80 successful grasp poses. In cases where a group contains fewer than 80 successful examples, we duplicate the available examples until the total reaches 80. Thus, the total number of positive samples used for training is 32,000. 

\subsubsection{Training}
For training our Energy-Based Model (EBM), we used a batch size $\mathcal{N}$ = 64 with the Adam optimizer, setting the learning rate to $1 \times 10^{-3}$ and applying a weight decay of $1 \times 10^{-5}$. The model was trained for 25 epochs, and the resulting model was then used to compute the gradient for SVGD using PyTorch's autograd. Our primary goal is not to achieve the best performance from the EBM but rather to demonstrate how the integration of a data-driven approach with optimization can effectively overcome the limitations inherent to each method.

\subsection{SVGD Optimization}\label{sec:SVGD}
In our work, we adapt SVGD within the Annealed Stein ICP (AS-ICP) framework~\cite{AS-ICP} for robust grasp synthesis. In AS-ICP ~\cite{SteinICP,AS-ICP}, the gradient \(\nabla_{\theta}\log p(\theta)\) in Equation~\eqref{eq:svgd} is replaced by the gradient of the SGD-ICP cost function with respect to the transformation parameters~\cite{sgdicp,sgdicp2}:

\begin{equation}
\nabla \log p(\theta) \approx -\gamma(t)\left(N\,\bar{g}(\theta, \mathcal{M}) + \nabla_{\theta}\log p(\theta)\right), 
\label{eq:c5:svgd-icp}
\end{equation}
where $\bar{g}(\theta, \mathcal{M})$ denotes the averaged gradient of the ICP cost function over a mini-batch $\mathcal{M}$, $N$ is a normalization factor equal to the number of particles, and $\gamma(t)$ is the annealing schedule. The prior gradient term, $\nabla_{\theta}\log p(\theta)$, is computed using Gaussian priors for translation and von Mises priors for rotation. However, in practical grasping scenarios, assuming a uniform prior over grasp poses fails to improve optimization performance, as it does not reflect the inherent structure of feasible grasps. To overcome this limitation, we propose augmenting the Stein ICP framework with a learned model that provides a more informative prior. Specifically, we replace the prior gradient $\nabla_{\theta}\log p(\theta)$ in Equation~\eqref{eq:c5:svgd-icp} with the gradient derived from an Energy-Based Model (EBM), denoted by $\nabla_{\theta} E(y,x)$:

\begin{equation}
\nabla \log p(\theta) \approx -\gamma(t)\left(w \times N\,\bar{g}(\theta, \mathcal{M}) + \nabla_{\theta} E(y,x)\right). 
\label{eq:c5:svgd-ebm}
\end{equation}

We observed that the magnitudes of the learned EBM gradient $\nabla_{\theta} E(y,x)$ and the matching gradient $N\bar{g}(\theta, \mathcal{M})$ differ significantly. This discrepancy can cause imbalanced updates during optimization, potentially degrading performance. To address this issue, we introduce a dynamic weighting factor $w$:

\begin{equation}
w = \frac{\| \nabla_{\theta} E(y,x) \|}{\|N\bar{g}(\theta_k , \mathcal{M})\|}.
\label{eq:c5:weight}
\end{equation}

We use the RBF kernel for translations and the dot product kernel for rotations. For the dot product kernel, the bandwidth parameter was set using the median heuristic~\cite{median}. However, in our experiments, we found that setting a fixed kernel bandwidth of $\sigma=3$ for translation consistently yielded better results than the median heuristic. 

We utilize the Adam optimizer with a learning rate of $1 \times 10^{-2}$ for SVGD parameter updates. Although the optimization requires a large number of iterations to converge, further increases in the learning rate result in parameter divergence and numerical instability. Additionally, we note that the ICP matching gradient provides strong guidance for translation parameters when the gripper is distant from the object, but its magnitude diminishes significantly as the gripper approaches the object's surface.

To take advantage of this behavior, we explicitly incorporate the matching gradient as a regularization term to accelerate early-stage translation optimization, gradually reducing its influence in later iterations. The final adaptive update rule for the transformation parameters is as follows:

\begin{equation}
\theta_{t+1} = \theta_t + \hat{\phi}_t^*(\theta_k) + \left(1-\frac{k}{K}\right)\bar{g}(\theta_k , \mathcal{M}),
\label{eq:c5:theta_update}
\end{equation}
where $\hat{\phi}_t^*(\theta_k)$ denotes the SVGD update, and the adaptive term $\left(1-\frac{k}{K}\right)$ progressively decreases the regularization effect of the matching gradient as the iteration index $k$ progresses toward the maximum iteration $K$.

At the end of the optimization process, we evaluate the matching error for each particle by selecting the minimum non-collision error across a predefined set of gripper preshapes, following the procedure described in~\cite{AS-ICP}. The complete optimization procedure is summarized in Algorithm~\ref{alg:sigsvgd}.

\begin{algorithm}[ht!]
    \scriptsize
    \caption{Stein Energy-Based Grasp}\label{alg:sigsvgd}
    \SetAlgoLined
    \DontPrintSemicolon
    \SetCommentSty{tiny}
    \KwIn{
        Point cloud of: 
        Gripper $\mathcal{S} = \{s_i\}_{i =1}^{N}$,
        Target Object $\mathcal{R} = \{ r_i\}_{i =1}^{M}$, 
        initial parameters ${\Theta}_0=\{\theta_0^j\}_{j=1}^{J}$,
        number of iteration $K$,
        SDF of gripper preshapes
    }
    \KwOut{
        $\theta$ that minimizes the sum of energy and matching error 
    }

    \While{k $\leq K$}
    {
        \For {each $\theta_k^j\in \Theta_k$ in parallel }
        {
            $\mathcal{S}_k^j \leftarrow$ Transform the source cloud with $\theta_k^j$\;
            $\nabla_{\theta} E(y,x)^j$ and $\bar{g}(\theta_k , M)^j \leftarrow$ Compute Stein variational gradient \eqref{eq:c5:svgd-ebm}\ then \eqref{eq:svgd}\;
            Collision avoidance using SDF\;
            Update $\theta$ \eqref{eq:c5:theta_update}\;
        }
        k = k + 1
    }
    \For{each $\theta^j \in \Theta_K$}
    {
        $\text{matching error}^j \leftarrow \min(\text{matching error}_{preshapes}^j)$\;
    }
    \Return $\theta = \operatorname*{argmin}_{\theta^j}\mathcal{(\mathrm{norm}(\text{energy}) + \mathrm{norm}(\text{matching error}))}$      
\end{algorithm}

\section{Experiments} \label{sec:Experiment}
\subsection{Ablation}\label{sec:ablation}
\subsection*{Data Processing}
\begin{figure}[t]
    \centering
    \small
    \includegraphics[width=6cm]{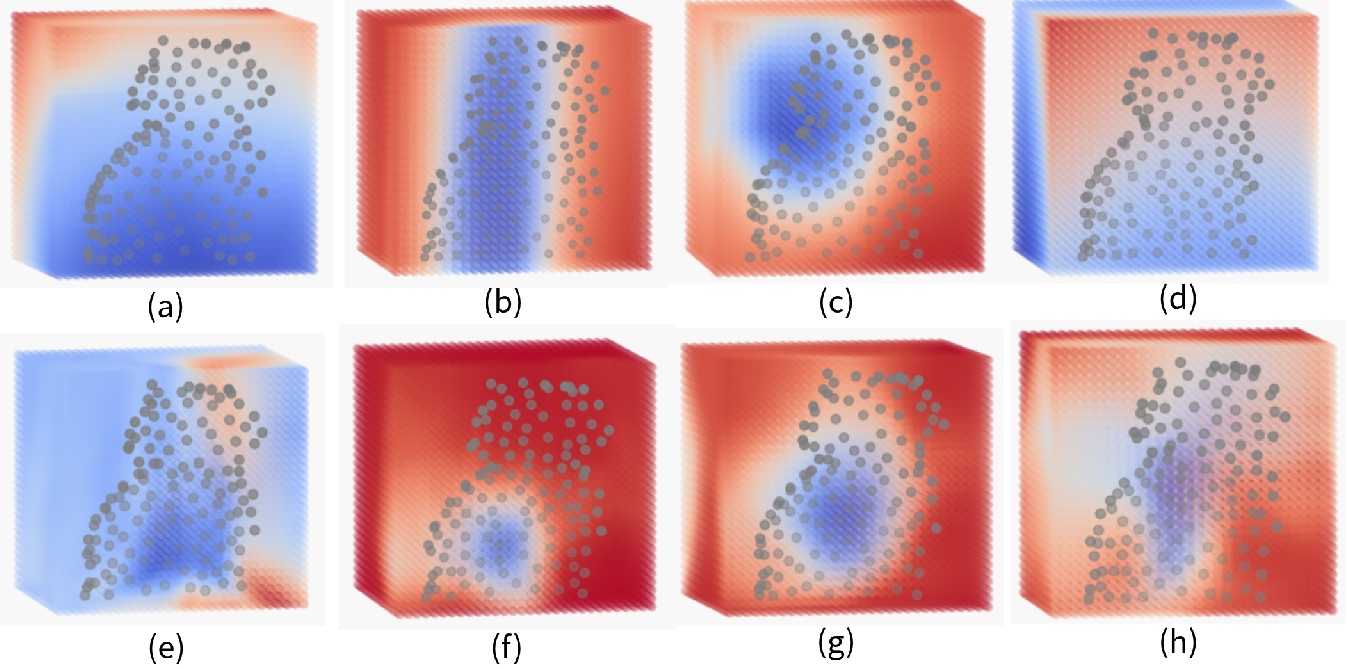}
    \caption{Energy landscapes for a top-down grasp under different training schemes. Blue indicates low energy (favorable grasps) and red indicates high energy (unfavorable grasps). Panels (a,e) use the full dataset; (b,f) use only positives with an equal number of negatives; (c,g) use negatives generated by uniform sampling; and (d,h) balance positives across groups (object orientation/elevation), yielding the best results. The top row excludes the gripper point cloud, while the bottom row includes it, leading to markedly improved performance.}
    \label{fig:energies}
\end{figure}

\textbf{Raw Data}. Training on the imbalanced AS-ICP dataset (28,555 positive vs. 72,461 total) limited the effectiveness of the contrastive loss, yielding poor separation between positive and negative examples (Figures \ref{fig:energies}(a), (e)). Balancing positives with an equal number of random negatives improved balance but flattened the energy field, making the model object-invariant (Figures \ref{fig:energies}(b), (f)).

\textbf{Localized Negative Sampling}. Generating negatives around each positive improved separation and introduced collision cases absent from the raw dataset. However, vertical grasp positions were poorly captured due to bias in the AS-ICP data (Figures \ref{fig:energies}(c), (g)).

\textbf{Group-based Sampling}. Partitioning data by orientation and elevation, balancing positives per group, and sampling localized negatives produced the best energy distributions (Figures \ref{fig:energies}(d), (h)).

Including the gripper point cloud as input further improved results (bottom row of Figure \ref{fig:energies}). However, even in the best configuration, energy minima were biased downward, reflecting dataset limitations—particularly the underrepresentation of top-down grasps and grasps near the table surface.

Overall, these results highlight that dataset distribution matters more than dataset size: well-structured training data yields a more accurate and reliable energy landscape, even with fewer examples.

\subsection*{Hybrid Model}
\begin{figure}[t]
    \centering
    \small
    \includegraphics[width=8cm]{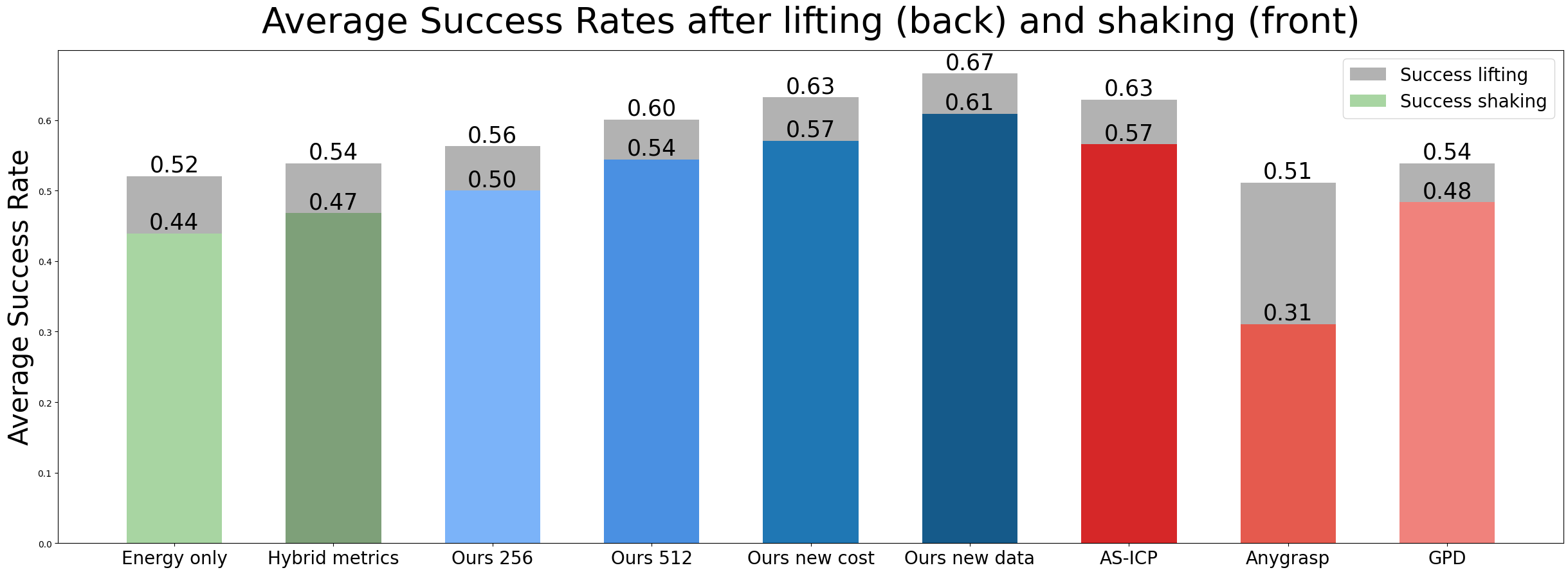}
    \caption{Plots of average success rate after lifting and shaking. The results demonstrate a clear trend of increasing performance as the model improves. Our method outperforms individual analytical and data-driven methods, other hybrid approaches, and baseline methods.}
    \label{fig:c5:success_2}
\end{figure}

\begin{figure}[t]
    \centering
    \small
    \includegraphics[width=8cm]{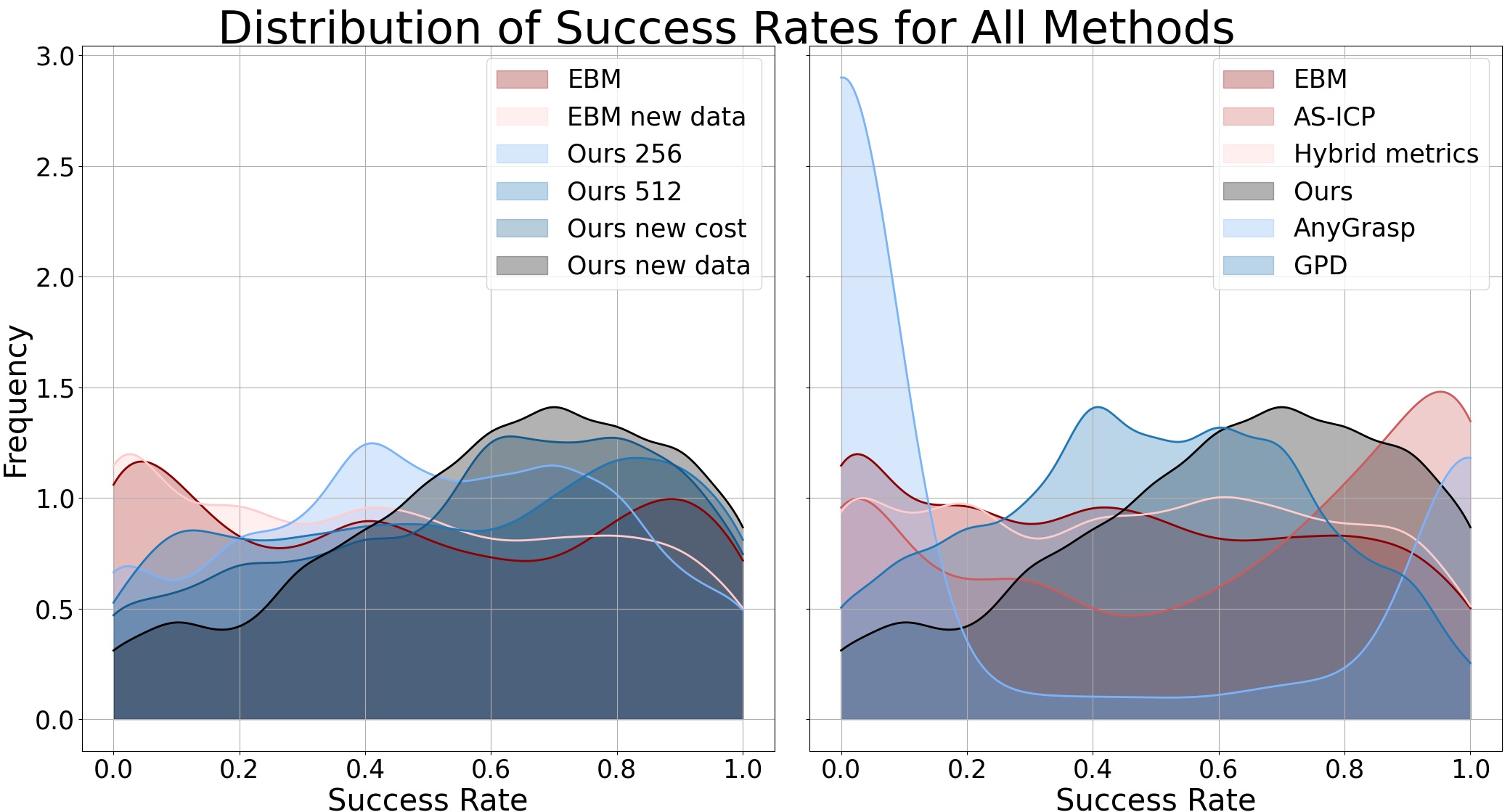}
    \caption{Kernel density estimates (KDEs) of success rates for all evaluated methods. The KDEs transform discrete success rate measurements into smooth, continuous probability distributions. \textbf{Left:} Hybrid model variants with different architectures and training data. \textbf{Right:} Baselines including analytical, learning-based, and hybrid approaches. Our model achieves sharper, higher-density peaks near high success rates, indicating improved reliability over baselines.}
    \label{fig:c5:distribution}
\end{figure}

\begin{table}[ht]
    \centering
    \scriptsize
    \setlength{\tabcolsep}{2.5pt} 
    \begin{tabular}{lccccc}
    \toprule
    Method           & Mean~($\uparrow$) & Std~($\downarrow$) & $\geq$ 0.1~($\uparrow$) & $\geq$ 0.5~($\uparrow$) & $\geq$ 0.9~($\uparrow$) \\
    \midrule
    Energy only      & 0.439 & 0.318 & 0.756 & 0.381 & 0.056 \\
    Hybrid metrics   & 0.468 & 0.312 & 0.795 & 0.431 & 0.054 \\
    Ours 256         & 0.500 & 0.285 & 0.864 & 0.451 & 0.057 \\
    Ours 512         & 0.544 & 0.310 & 0.851 & 0.513 & 0.093 \\
    Ours new cost    & 0.571 & 0.291 & 0.890 & 0.586 & 0.082 \\
    Ours new data    & \textbf{0.609} & 0.270 & \textbf{0.920} & \textbf{0.623} & 0.095 \\
    AS-ICP           & 0.566 & 0.367 & 0.784 & 0.567 & 0.192 \\
    AnyGrasp         & 0.311 & 0.440 & 0.347 & 0.306 & \textbf{0.254} \\
    GPD              & 0.484 & \textbf{0.259} & 0.873 & 0.424 & 0.022 \\
    \bottomrule
    \end{tabular}
    \caption{Comparison of methods by group-level mean, standard deviation, and fraction of groups exceeding selected success thresholds. 
    The arrows indicate whether higher ($\uparrow$) or lower ($\downarrow$) values are preferred. 
    Thresholds (0.1, 0.5, 0.9) denote the fraction of groups whose success rate is above 10\%, 50\%, and 90\%, respectively, providing a sense of performance across low, moderate, and high success regimes.}
    \label{tab:c5:compare}
\end{table}

The main focus of this work is to introduce and evaluate a hybrid optimization framework for grasp synthesis. While ICP and EBM are used as representative analytic and learning-based components, the framework is modular and can be extended to other methods, though a full exploration of alternatives is beyond the present scope. The aim is not to train the single best EBM, but to demonstrate that as the learned model improves, grasp performance correspondingly increases.

To evaluate the framework, we conduct ablation studies and systematically compare it against analytical (ICP), learning-based (EBM, GPD, AnyGrasp), and hybrid baselines (Hybrid metrics). A common strategy is to apply ICP as post-processing to learning-based grasps. However, as shown in~\cite{AS-ICP}, if the initial grasps are poor, ICP refinement provides little improvement.

Our ablations examine both model architecture and dataset composition.

\begin{itemize}
    \item \textbf{Dataset 1:} 32,000 positive examples from AS-ICP. Models include:
    \begin{itemize}
        \item \textbf{EBM:} Baseline (256-dim encoder, energy only). 
        \item \textbf{Ours 256:} Adds matching error.
        \item \textbf{Ours new cost:} Adds two extra loss terms.
        \item \textbf{Ours 512:} Larger encoder (512-dim) with combined point cloud features.
    \end{itemize}

    \item \textbf{Dataset 2:} 3,280 positives generated by “Ours new cost”. Models include:
    \begin{itemize}
        \item \textbf{EBM new data:} Energy only (256-dim).
        \item \textbf{Ours new data:} Energy  (256-dim) + matching error + additional losses.
    \end{itemize}
\end{itemize}

The additional loss terms used for Ours new cost are defined as follows:

\begin{equation}
\mathrm{PartialError}(R_i) = \mathrm{sigmoid} \left( 10 \left( \left| \left( R_i \mathbf{e}_z \right)_y \right| - 0.5 \right) \right),
\end{equation}
where $\mathrm{sigmoid}(x) = \frac{1}{1 + e^{-x}}$, $R_i$ is the $i$-th rotation matrix, and $\mathbf{e}_z = [0, 0, 1]^T$, which penalizes near-horizontal grasps prone to unseen collisions.

\begin{equation}
\mathrm{PointInGrasp} = \exp\left( -\frac{1}{10} \sum_{i=1}^N
\big( p_i \in \mathcal{B} \big)
\right),
\end{equation}
where $\mathcal{B}$ is the volume within the grasp (i.e., the region enclosed by the gripper), which encourages object points within the gripper volume.

Figure~\ref{fig:c5:success_2} and Figure~\ref{fig:c5:distribution} (left) show progressive improvement: the raw EBM produces a wide spread of success rates, while adding costs and model capacity yields more peaked, consistent outcomes. Refining the dataset further boosts performance, with “Ours new data” achieving the highest density near high success rates.

By contrast, Figures~\ref{fig:c5:distribution} (right) benchmarks performance against baselines. AS-ICP and AnyGrasp show extreme outcomes; GPD produces a spread closer to our hybrids. Overall, our framework yields higher mean success and tighter distributions, though—as Table~\ref{tab:c5:compare} shows—it produces fewer near perfect (90\%) cases, reflecting the stochasticity of learned components.

\begin{figure*}[t]
    \centering
    \includegraphics[width=12cm]{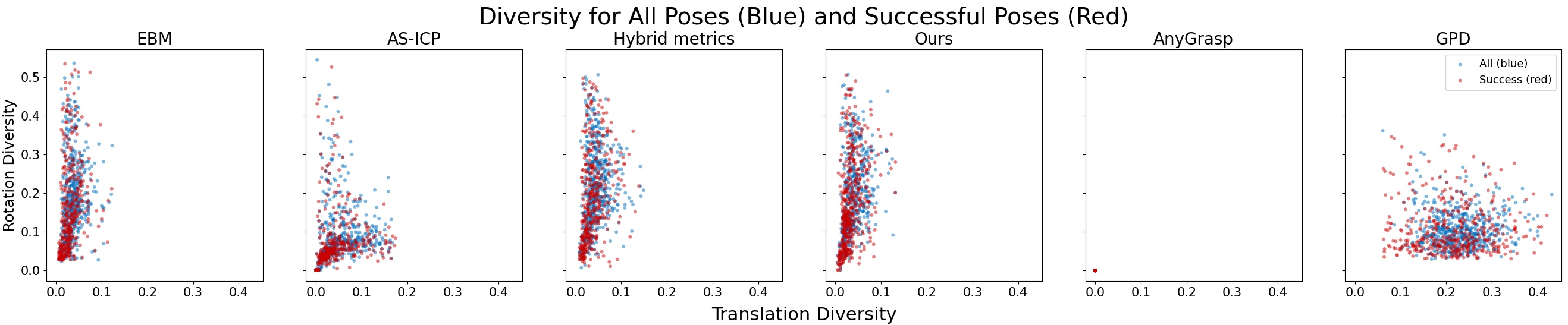}
    \includegraphics[width=12cm]{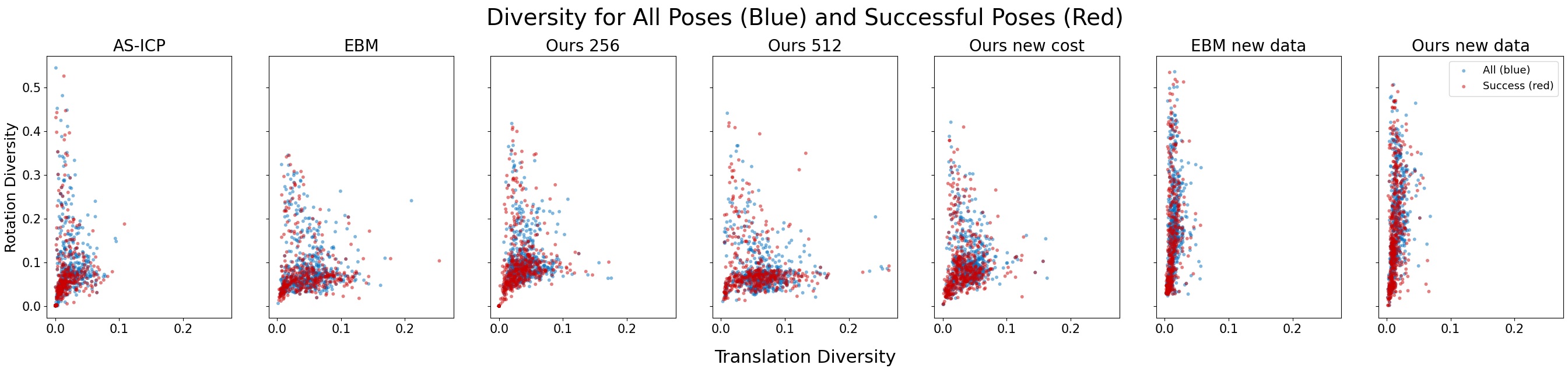}
    \caption{Comparison of grasp diversity across baseline and hybrid methods. Translation diversity (x-axis, up to 98th percentile) and rotation diversity (y-axis) are shown for all attempted (blue) and successful (red) grasps. Among baselines, AnyGrasp produces identical poses, GPD yields high translation but low rotation diversity, AS-ICP concentrates at low diversity, and EBM shows high rotational diversity. Hybrid models generally mirror their underlying EBM: the initial dataset leads to high translation diversity, while the refined dataset reduces translation but substantially increases rotational diversity.}
    \label{fig:c5:diversity_models}
\end{figure*}

In Figure~\ref{fig:c5:diversity_models}, grasp diversity is plotted in terms of translation (x-axis) and rotation (y-axis), computed from pairwise distances between successful grasps. Translation is measured by standard deviation, while rotation differences are weighted more heavily for nearby poses. Because results are drawn from the best of ten trials, these plots reflect repeatability.

Among baselines (top), GPD exhibits high translation but low rotation diversity, AnyGrasp produces nearly identical poses, and AS-ICP clusters at low diversity. EBMs spread more broadly, especially in rotation. Hybrid models (bottom) inherit the behavior of their learned components: the initial EBM-trained model resembles ICP, while the refined “Ours new data” variant shows lower translation but higher rotation diversity, yielding consistent contact points with flexible approach angles.

Overall, dataset refinement and model design improve repeatability, whereas purely analytic or sampling-based methods lack consistency, and generative models risk failure by producing identical grasps. The hybrid framework achieves a favorable balance of stability and adaptability.

\subsection{Simulation}
\begin{figure*}[t]
    \centering
    \small
    \includegraphics[width=14cm]{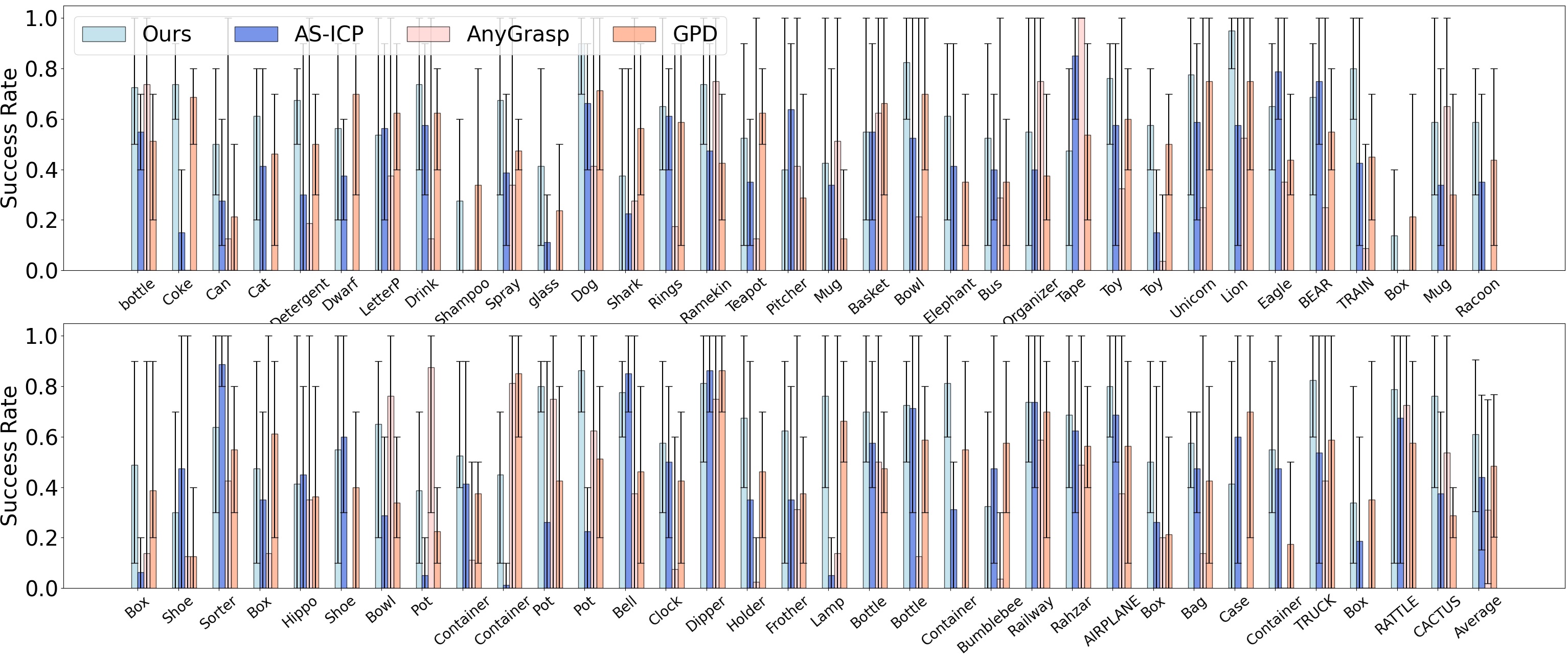}
    \caption{Simulation results comparing our approach with baseline methods. Our method achieved an average success rate of 60.9\%, outperforming AnyGrasp (31.1\%), GPD (48.4\%), and AS-ICP (56.6\%). We also achieved higher object-based (not group-based) minimum and maximum success rates, as indicated by the error bars.}
    \label{fig:c5:simulation}
\end{figure*}

\begin{figure}[t]
    \centering
    \small
    \includegraphics[width=5.5cm]{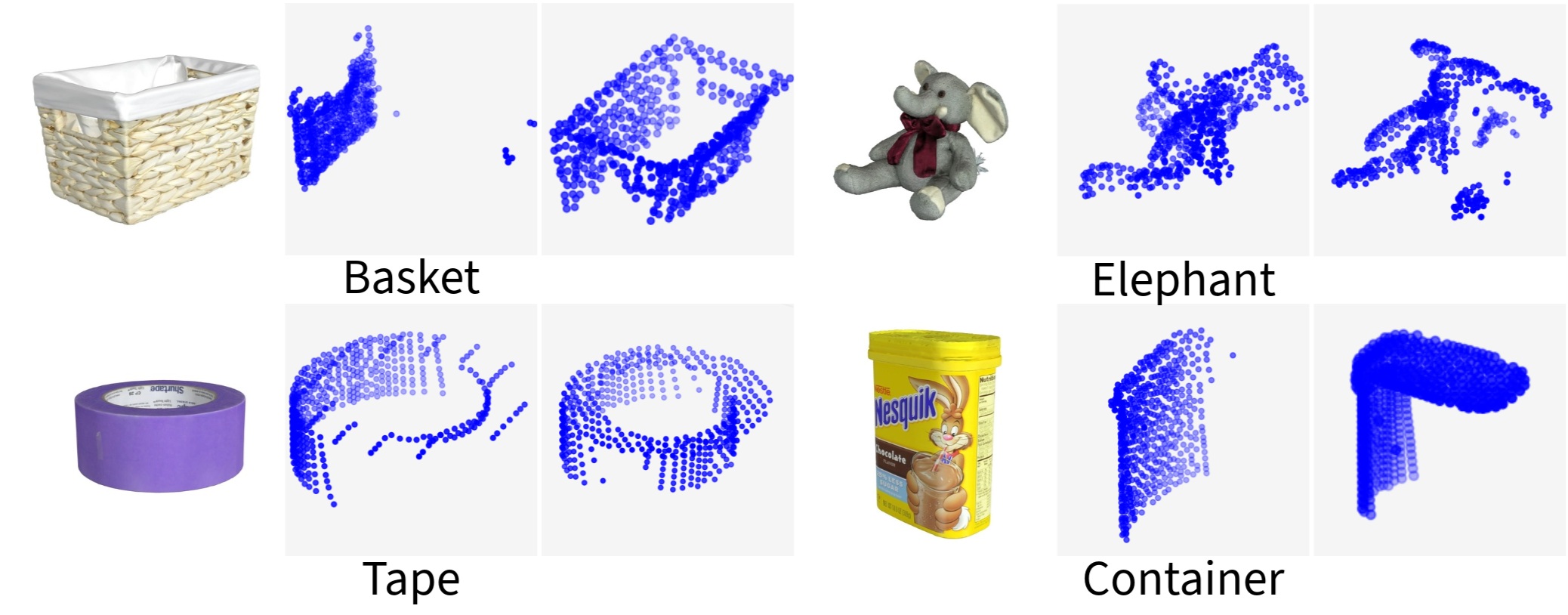}
    \caption{Examples of objects and their corresponding point clouds.}
    \label{fig:c5:pcd}
\end{figure}

\begin{figure}[t]
    \centering
    \small
    \includegraphics[width=5.5cm]{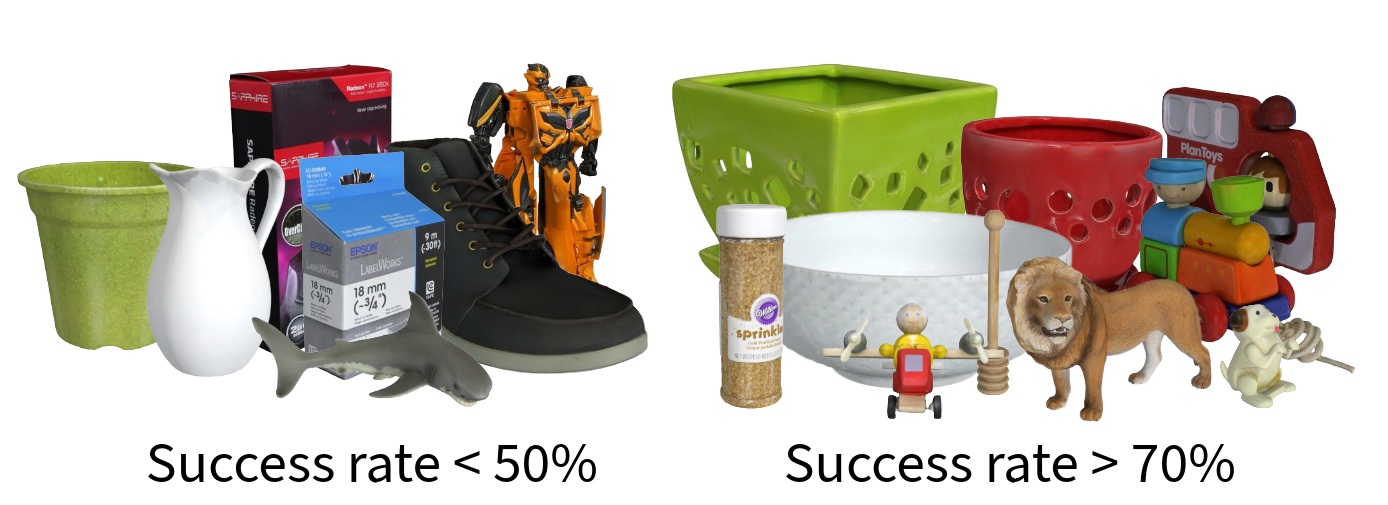}
    \caption{Examples of objects used in simulation for which our method achieves less than 50\% success rate (left) and more than 70\% (right). There is no obvious visual distinction between object categories with less than $50\%$ success rate and those above $70\%$, suggesting that occlusion and viewpoint-specific visibility are key factors.}
    \label{fig:c5:obj}
\end{figure}

We selected 67 objects from the Google Scanned Objects dataset and captured their point clouds in the Isaac Gym simulator~\cite{Isaac}. Some objects were rescaled to ensure that the Franka robotic arm could grasp them securely—large enough for a stable hold, yet not so small as to fit entirely within the gripper. All training and optimization experiments were conducted on a laptop equipped with an RTX 2070 GPU. The grasp pose’s origin is aligned with the known object origin. In real experiments, the origin is approximated by the centroid of the segmented object point cloud, computed from a stable reference view to avoid inconsistencies caused by occlusions and changing viewpoints.

For evaluation, we compared our approach with three baselines—AnyGrasp~\cite{AnyGrasp}, Grasp Pose Detection (GPD)~\cite{GPD}, and AS-ICP~\cite{AS-ICP}—all of which generate grasps from partial point clouds. For each method, a set of candidate grasps was generated, the top-scoring pose was executed, and this was repeated ten times to compute the average success rate. 

Simulation results for the 67 objects are shown in Figure~\ref{fig:c5:simulation}. Our method achieved an average success rate of $60.9\%$ over 5,360 grasps, outperforming AnyGrasp ($31.1\%$), GPD ($48.4\%$), and AS-ICP ($56.6\%$). We also achieved higher object-based mean minimum and maximum success rates, as indicated by the error bars. This differs from our previous analysis, which was group-based. However, conventional object-level analysis is less meaningful in our setting, since partial point clouds from different viewpoints can differ significantly (Figure~\ref{fig:c5:pcd}). For example, Figure~\ref{fig:c5:obj} shows that there is no obvious visual distinction between object categories with less than $50\%$ success rate and those above $70\%$, suggesting that occlusion and viewpoint-specific visibility are key factors.

Our approach also demonstrates strong generalization ability: although the model was trained on only the first 41 objects, it achieved a higher success rate on unseen objects ($64.6\%$) than on the training set ($58.6\%$).

\begin{table}[ht]
\centering
\scriptsize
\begin{tabular}{l c}
\toprule
\textbf{Ours (Franka)} & \textbf{Time (s)} \\
SVGD with ICP only & 0.653 \\
Initial EBM loading  & 1.17 \\
Energy and gradient computation & 0.458 \\
Overall & 2.28 \\
\midrule
\textbf{AS-ICP (Franka)} & \textbf{Time (s)} \\
SVGD followed by SGD & 2.42 \\
\bottomrule
\end{tabular}
\caption{Computation time breakdown for the hybrid method compared to original AS-ICP.}
\label{c5:tab:comp_time}
\end{table}

As shown in Table~\ref{c5:tab:comp_time}, the use of a single preshape during optimization significantly reduces ICP computation time (0.653 s) compared to AS-ICP (2.42 s), which used 7 preshapes to ensure a high success rate. The complete hybrid optimization process (including model loading time and optimization time) requires approximately 2.28s, demonstrating that it is faster than the AS-ICP method even when including the overhead of loading the EBM.

\subsection{Real Experiment}
\begin{table*}[t]
\scriptsize
\setlength{\tabcolsep}{3pt}
\centering
\begin{tabular}{|l|c|c|c|c|c|c|c|c|c|c|c|}
\hline
\textbf{Object} & Detergent & Washing Liquid & Spray & Holder & Bag & Mouse & Glass Container & Box & Water Bottle & Hand Help Tool & \textbf{Avg.} \\
\hline
\textbf{AS-ICP} & 0.80 & 1.00 & 0.80 & 0.60 & 0.40 & 0.20 & 0.60 & 0.40 & 0.80 & 0.80 & \textbf{0.64} \\
\hline
\textbf{Ours} & 0.80 & 1.00 & 1.00 & 0.60 & 0.60 & 0.40 & 0.60 & 1.00 & 1.00 & 0.80 & \textbf{0.78} \\
\hline
\end{tabular}
\caption{Comparison of average success rates per object between AS-ICP and our method.}
\label{tab:real_comparison}
\end{table*}

\begin{figure}[t]
    \centering
    \small
    \includegraphics[width=7cm]{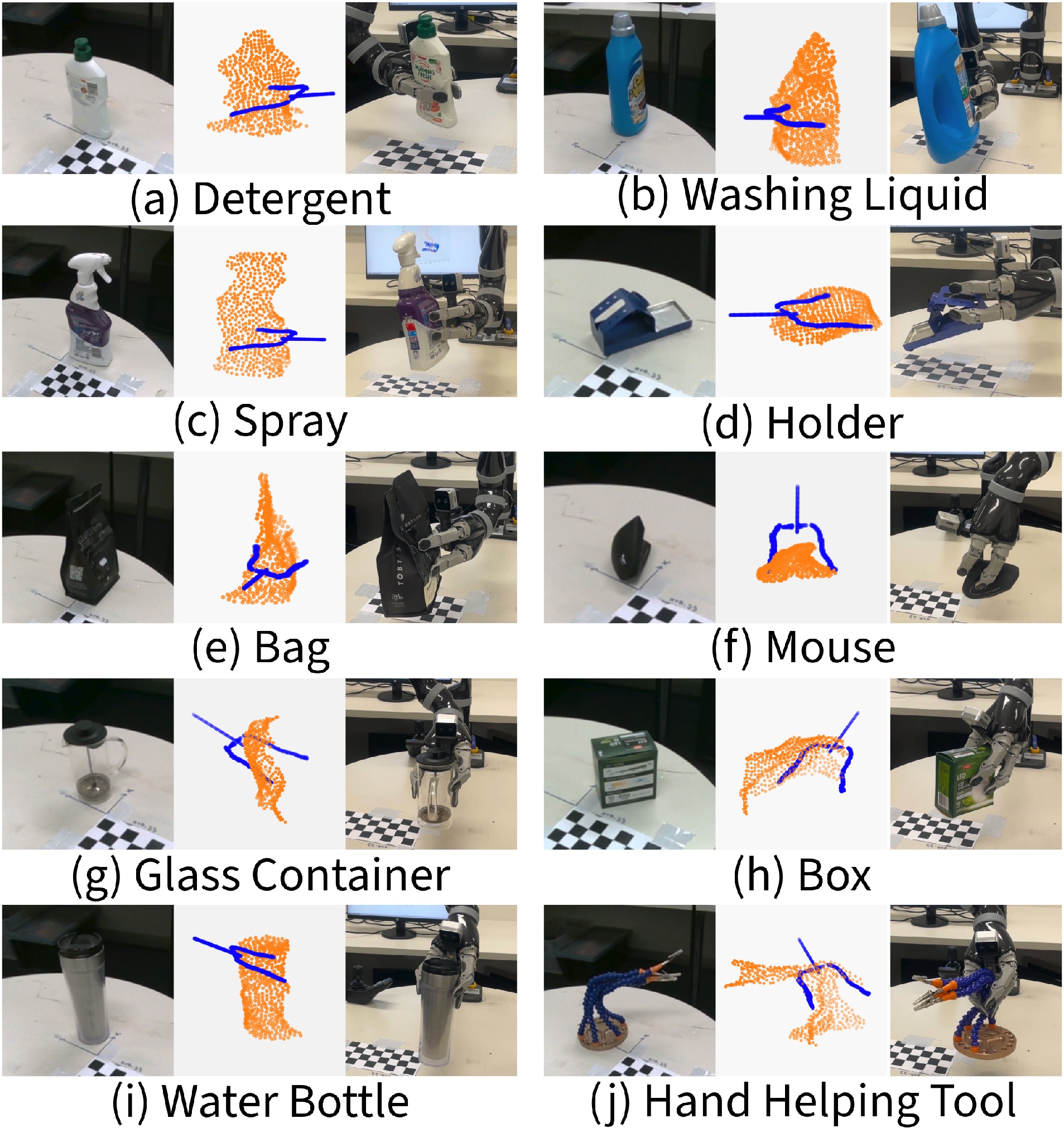}
    \caption{Illustration of the real experiment. For each object: left—camera view, middle—scanned point cloud with predicted grasp, right—KG3 gripper executing the grasp.}
    \label{fig:c5:real}
\end{figure}

To validate our algorithm, we conducted experiments using a KG3 gripper mounted on a Kinova arm. We selected ten everyday objects and placed each in five different orientations, capturing the point clouds with a wrist-mounted camera. Overall, we achieved a 78\% success rate across fifty grasps, as summarized in Table~\ref{tab:real_comparison}. Figure~\ref{fig:c5:real} displays a single view of each object along with the corresponding point clouds and generated grasps.

Our algorithm is robust to noisy and partial point clouds. However, it struggles when the observed point cloud deviates significantly from the true object geometry—for instance, with the Holder (d) and Mouse (f), where table noise reduces accuracy, and the transparent Glass Container (g), where poor sensing leads to failures. In contrast, the Box (h) shows clear improvement over AS-ICP, highlighting the method’s strength in reducing misalignment.

\section{Summary and Discussion} \label{sec:conclusion}
Through ablation studies, we demonstrated the critical role of structured datasets in training EBMs and showed that the hybrid formulation consistently outperforms each component in isolation. Experimental results confirm that our method achieves high success rates on both seen and unseen objects.

Several promising directions emerge from this work. First, because AS-ICP is gripper-invariant, our hybrid framework inherits this property. Training the EBM on data from multiple grippers would enable grasp synthesis across different end-effectors without architectural changes. Second, our method is not a competitor to existing approaches, but a modular framework into which other learning-based components can be integrated—such as alternative grasp synthesis networks or other analytical metrics. Third, our pipeline—generating data through an analytical method (AS-ICP), trains an EBM, then hybrid optimization—mirrors, in simplified form, the broader paradigm of simulation-based training. This perspective exposes a deeper opportunity: if simulators themselves are collections of analytical methods, then our hybrid framework—where a learned model augments the very method that generated its training data—suggests a path toward closed-loop refinement between simulation, learning, and optimization.

%%%%%%%%%%%%%%%%%%%%%%%%%%%%%%%%%%%%%%%%%%%%%%%%%%%%%%%%%%%%%%%%%%%%%%%%%%%%%%%%

%\addtolength{\textheight}{-9cm}

%%%%%%%%%%%%%%%%%%%%%%%%%%%%%%%%%%%%%%%%%%%%%%%%%%%%%%%%%%%%%%%%%%%%%%%%%%%%%%%%

\bibliographystyle{ieeetr}
\bibliography{references}

\end{document}